\documentclass{article}
\usepackage{natbib,alifeconf}

\usepackage{algorithm}
\usepackage{algpseudocode}
\usepackage[euler]{textgreek}
\usepackage{graphicx}
\usepackage{setspace}
\usepackage{subeqn}
\usepackage{subcaption}
\usepackage{url}

\title{Evolutionary Robotics on the Web with WebGL and Javascript}

\author{Jared M. Moore, Anthony J. Clark \and Philip K. McKinley \\
Dept. of Computer Science and Engineering \\
Michigan State University, East Lansing, MI, USA 48824 \\
moore112@msu.edu}

\begin{document}

\maketitle

\paragraph{Introduction.}

Web-based applications are highly accessible to users, providing rich, interactive content while eliminating the need to install software locally.  
Previous online evolutionary demonstrations~(PicBreeder~\citep{Secretan2008}, Endless Forms~\citep{Clune2011}, Ludobots~\citep{Bongard2012}, and BoxCar2D) have successfully demonstrated concepts to a broad audience.
However, evolutionary robotics~(ER) has faced challenges in this domain as web-based technologies have not been amenable to 3D physics simulations.  
%
%
Traditionally, physics-based simulations require a local installation and a high degree of user knowledge to configure an environment, but the emergence of Javascript-based physics engines enables complex simulations to be executed in web browsers.  
These developments create opportunities for ER research to reach new audiences by increasing accessibility.  

In this work, we introduce two web-based tools we have built to facilitate the exchange of ideas with other researchers as well as outreach to K-12 students and the general public.  
The first tool is intended to distribute and exchange ER research results, while the second is a completely browser-based implementation of an ER environment.  
We use WebGL~\citep{WebGL}, ThreeJS~\citep{ThreeJS}, and PhysiJS~\citep{PhysiJS} to build online ER applications.  


\paragraph{WebGL-Based Visualizer.}

Visualization is a powerful means to communicate concepts and discoveries.  
Many researchers in ER and Artificial Life produce videos of simulations to disseminate results from their research.
However, videos are usually limited to what the author selects through specific angles or edits.  
Our first tool is a WebGL-based visualization system that provides more capability to the viewers.  
Rather than passively watching a video, viewers can interact with the results, exploring from new angles and focal points.  
A demonstration of this work can be seen at: \url{http://jaredmmoore.com/WebGL_Visualizer/visualizer.html}.  

\begin{figure}[htb!]
\includegraphics[width=3.25in]{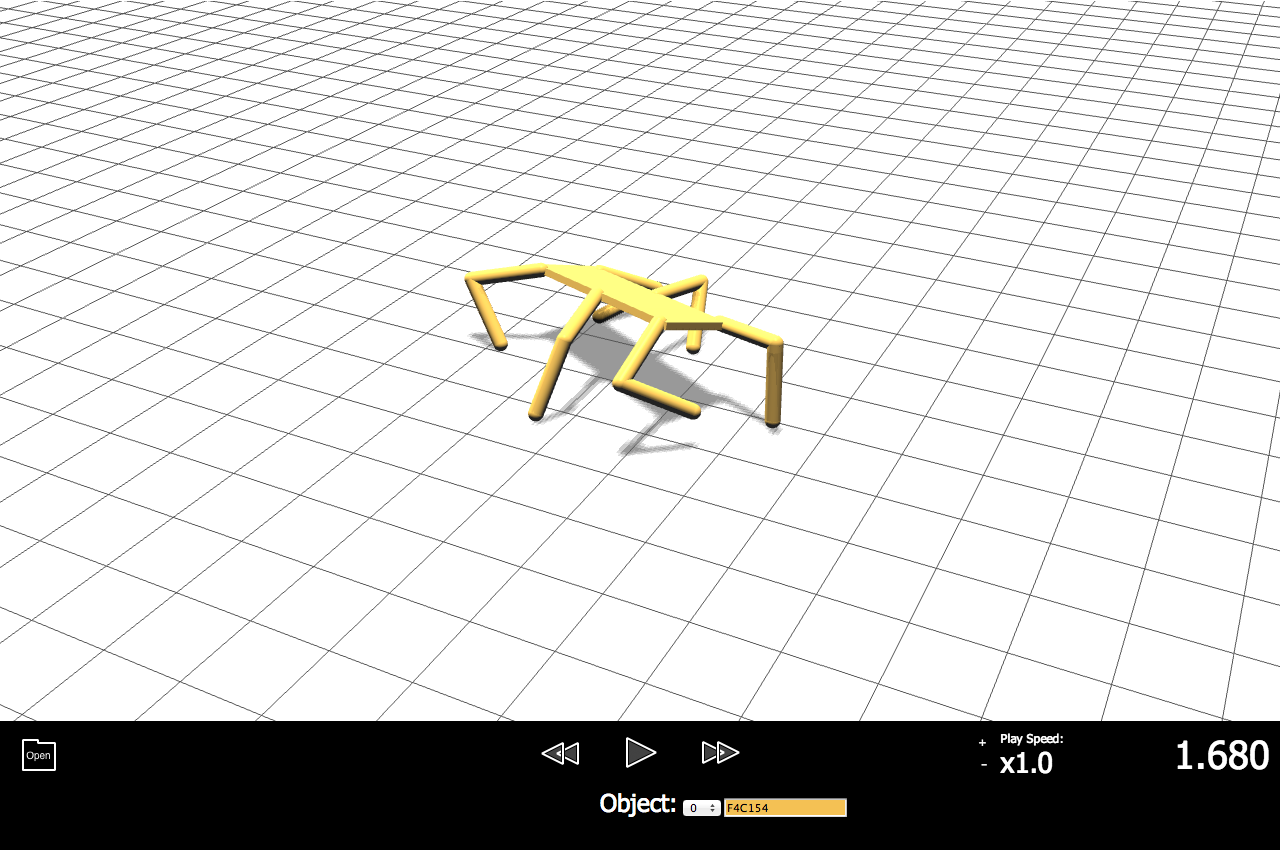}
\caption{WebGL allows for browser-based graphics based on the OpenGL ES 2.0 Standard.  Here, an evolved hexapod robot is shown in the WebGL Visualizer.  Viewers can change the camera's location, alter the color of each body, and adjust playback speed to customize their experience.}
\label{fig:webgl_visualizer}
\end{figure}

In addition to enhanced interaction with results, the WebGL-based visualizer decouples the physics engine and visualization.  
Researchers often employ different engines (e.g. ODE, Bullet, or PhysX), each of which may require a custom visualization to be written.  
Instead, our proposed visualizer uses a common log file standard which includes the initial simulation configuration and state of the bodies at each timestep.  
This eliminates the need to reimplement a visualization engine, allowing a researcher to focus on developing simulations and results.  
Moreover, results can easily be shared by sending the log file, as opposed to installing a complete physics simulation environment.  
The visualization engine is strictly for playback, a true online environment would allow for interaction and physics simulation.  


\paragraph{Evolve-a-Robot.}

Modern Javascript has led to an increase in browser functionality, extending web pages from static information to interactive applications.  
Accordingly, the implementation of 3D physics engines in Javscript allows for online ER environments that does not require the installation of custom software.  
A user can visit a web page with a modern browser and get started immediately.  
Browser-based physics engines have previously been employed in BoxCar2D~(\url{http://www.boxcar2d.com}), demonstrating genetic algorithm concepts through an interactive application in which users can evolve cars.  
\vspace{0.1in}

\begin{figure}[htb!]
\includegraphics[width=3.25in]{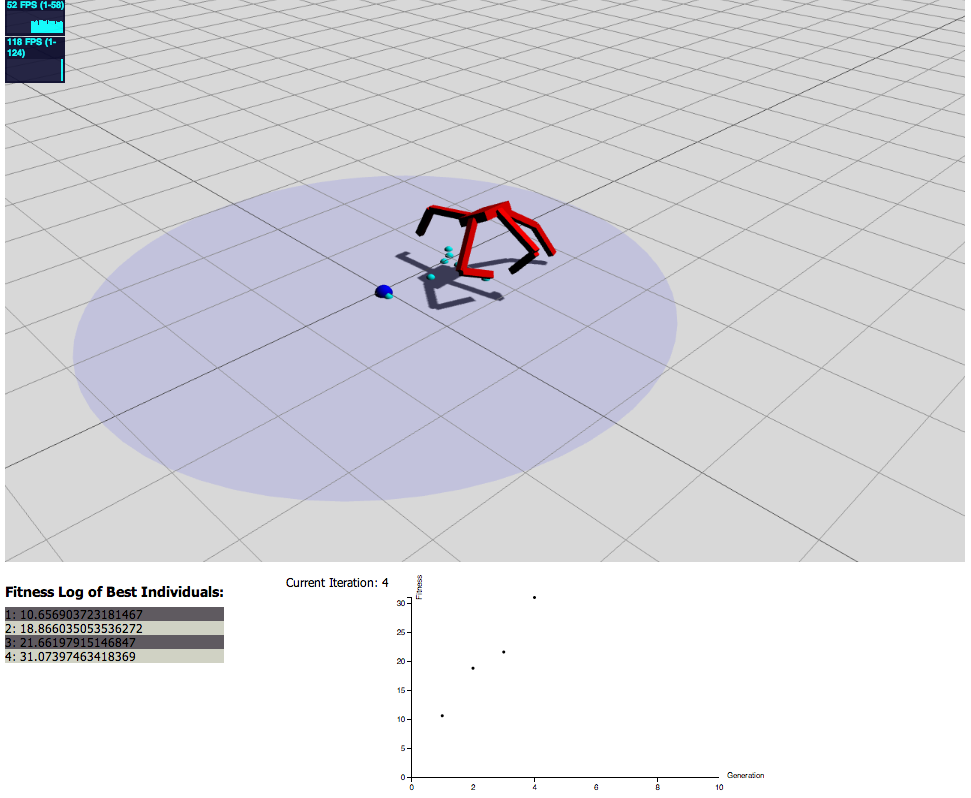}
\caption{Evolve-a-Robot uses the PhysiJS port of Bullet to deliver an ER environment completely in a browser.  Users can choose the animat and watch as a 1+1 algorithm evolves animats over time.}
\label{fig:evorobo}
\end{figure}

Until recently however, browser-based 3D physics engines had not reached the level of maturity needed to construct a full simulation environment.  
Our ER environment employs PhysiJS~\citep{PhysiJS}, a port of the Bullet physics engine~\citep{Bullet}, to handle real-time physics simulation.  
Figure~\ref{fig:evorobo} shows a screen capture of the application; a live demo can be found at \url{http://jaredmmoore.com/EvoEnv/evo_main.html}.
Our current environment features a choice of three different animats (a quadruped, an octopod, and a Karl Sims inspired animat~\citep{Sims1994}) evolved with a 1+1 algorithm.  
The evolutionary system includes the ability to revisit the best solutions, and a scatter plot of fitnesses.  
By leveraging the D3.js library, we are able to update the scatter plot dynamically providing a visual representation of evolution for a viewer.  

In the current prototype, the evolutionary parameters are fixed.  
Future versions will add them as user defined values.  
Moreover, the application will be more interactive, allowing users to alter values such as mutation rate and immediately view the resulting changes in the simulation.  
In addition, we plan to develop a distributed GA implementation.  
Users will load the web page and obtain a genome from the server.  
Simulations will be carried out locally, with results communicated back to the server.  

\vspace{0.2in}
\paragraph{Summary.}

Web-based technologies are opening new avenues for conducting science and outreach.  
Leveraging Javascript and WebGL, we have created two tools to assist researchers in visualizing results and communicating evolutionary concepts with other researchers, K-12 students, and the general public.  
In future work, we plan to extend both applications by adding features to increase user interaction.  

\vspace{-0.1in}
\paragraph{Source Code.}

Source code for the two tools is available at:

WebGL-Based Visualizer: \\
\url{https://github.com/jaredmoore/WebGLVisualizer}

Evolve-a-Robot: \\
\url{https://github.com/jaredmoore/EvolveARobot}

\vspace{-0.1in}
\paragraph{Acknowledgements}
The authors gratefully acknowledge the contributions of Brian Connelly and the support of the BEACON Center at Michigan State University. 
This work was supported in part by National Science Foundation grants CNS-1059373, CNS-0915855, and DBI-0939454, and by a grant from Michigan State University.

\bibliographystyle{plainnat}
\bibliography{References/references}

\end{document}